\documentclass[11pt,journal,onecolumn]{IEEEtran}
\ifdefined\pdfminorversion\pdfminorversion=7\fi
\usepackage{cite}

\usepackage[T1]{fontenc}
\usepackage[table]{xcolor}
\usepackage{graphicx}
\usepackage[hyphens]{url}
\usepackage{amsmath,amssymb,bm}
\usepackage{booktabs,multirow,array}
\usepackage{placeins}
\usepackage{xspace}
\newcommand{\ie}{i.e.\xspace}
\usepackage[hidelinks]{hyperref}

\definecolor{auxgray}{gray}{0.48}
\newcommand{\graytxt}[1]{{\color{auxgray}#1}}

\definecolor{revred}{RGB}{190,0,0}

\newif\ifshowchanges
\showchangesfalse

\newsavebox{\lodeottablebox}

\title{LoDEOT: Low-Dimensional and Efficient Offset Tokens
for Building Footprint Extraction from Off-Nadir Imagery}

\author{%
  Kai Li\textsuperscript{1,2}, Zigan Zhou\textsuperscript{1},
  Zhenyang Li\textsuperscript{3}, Hui Shan\textsuperscript{4},
  Zhe Chen\textsuperscript{5}\\[3pt]
  Yupeng Deng\textsuperscript{2}, Zhihao Xi\textsuperscript{2},
  Yu Meng\textsuperscript{2}, Yifan Peng\textsuperscript{3},
  Xiangyu Zhao\textsuperscript{1}\\[6pt]%
{\small
  \textsuperscript{1}City University of Hong Kong\\
  \textsuperscript{2}University of the Chinese Academy of Sciences\\
  \textsuperscript{3}University of Hong Kong\\
  \textsuperscript{4}Zhejiang University\\
  \textsuperscript{5}University of Southampton\par}%
}
\date{}
\hypersetup{%
  pdftitle={LoDEOT: Low-Dimensional and Efficient Offset Tokens for Building Footprint Extraction from Off-Nadir Imagery},
  pdfauthor={Kai Li, Zigan Zhou, Zhenyang Li, Hui Shan, Zhe Chen, Yupeng Deng, Zhihao Xi, Yu Meng, Yifan Peng, Xiangyu Zhao}%
}

\begin{document}

\maketitle
\pagestyle{plain}
\thispagestyle{plain}

\begin{abstract}
    Instance-level roof-to-footprint offset (RFO) prediction is central to extracting building footprints from off-nadir imagery. Query-based pipelines commonly use high-dimensional instance tokens to predict signed two-dimensional RFOs. We investigate whether RFO prediction can instead use a compact offset token. Under local pinhole projection and vertical-extrusion assumptions, the idealized RFO map admits a five-parameter sufficient descriptor comprising intrinsic shape, composite amplitude, and relative geometry. This factorization provides a structural prior for a five-dimensional offset token, whose channels learn task-relevant latent representations through end-to-end training. Based on this design, we propose LoDEOT, which retains high-dimensional instance tokens for detection and segmentation but maps instance-token, concentration-gated roof, and box--mask evidence to a five-dimensional offset token followed by an independent two-dimensional readout. Known denoising-query target indices further align each supervised decoder-layer estimate with the same clean instance RFO, organizing successive predictions as target-aligned recovery under perturbed query conditions. Experiments on five real-world building datasets demonstrate the effectiveness of LoDEOT for building footprint extraction. Experiments on real-world building datasets demonstrate that a five-dimensional offset token can support accurate RFO prediction. On BONAI, LoDEOT achieves the best roof-detection bAP and bAP50 and leads all five offset-corrected footprint metrics among the evaluated end-to-end methods, with FAP50 of 54.58 and mEPE of 5.23 pixels. Its FAP50 exceeds those of the evaluated end-to-end baselines by 7.56--16.85 percentage points. The code of this work will be 
available at \url{https://github.com/ZiganZHOU/LoDEOT}.
\end{abstract}

\section{Introduction}

Accurate building footprints support urban mapping, population analysis,
disaster response, and 3D city modeling.
Raster segmentation and polygonal extraction improve building
delineation~\cite{maggiori_2017_semantic,demir_2018_deepglobe,ji_2019_fully,li_2021_joint,girard_2021_polygonal},
but viewing direction and building height displace visible roofs from
their ground outlines in off-nadir imagery.
The signed roof-to-footprint offset (RFO) describes this roof-to-footprint
displacement in the image plane; its accurate prediction supports
footprint localization, boundary alignment, height estimation, and
3D building reconstruction~\cite{wang_2023_learning,li_2024_3d}.

RFO prediction has progressed from convolutional roof--offset branches, including LOFT and MLS-BRN~\cite{wang_2023_learning,li_2024_3d}, to prompt-driven and token-based instance models~\cite{li_2024_promptdriven,li_2025_polyfootnet}. DG-BRF combines diffusion-based roof and facade segmentation with geometric priors for offset estimation~\cite{hu_2025_3d}. These methods improve observations, supervision, or output structure, yet the offset representation used to predict the offset itself remains largely unexamined, \textit{e.g.,} commonly used 256, 512, 1024 dimensions for token design~\cite{li_2024_promptdriven,li_2025_polyfootnet,li2025dragosm,li_2026_rbfsam}. A generic instance token must stay expressive for classification, localization, and segmentation; forcing the task-specific offset route to inherit the same width adds width-dependent offset-head parameters, while roof appearance, mask reliability, and box--mask asymmetry remain only implicitly available. The issue is not whether to compress the shared token, but whether the RFO route can be decoupled into a compact offset head. 

Three observations guide our design. First, extensive linear probing experiments~\cite{oquab2023dinov2,rombach2021geometry,zhou2026understanding,bahmani2025ac3d} have demonstrated that Transformer-based models implicitly encode 3D photographic information (e.g., camera intrinsics) during training. Second, under local pinhole-camera, vertical-extrusion, and visibility assumptions, principal-point cancellation, coordinate gauges, scale coupling, and consolidation of relative geometric factors yield an exact five-parameter sufficient description of the idealized RFO map. This factorization motivates a five-dimensional candidate for the offset head; it does not establish a unique, minimal, or universally optimal neural width. Third, successive offset decoder layers support target-aligned recovery of the offset tokens: the offset tokens are gradually close to the ground truth offset. This is a layer-by-layer denoising process. 


Guided by these observations and a first-principles analysis of the mathematical relationship between RFO and physical parameters, \ie, building height \& camera parameters, we propose an RFO prediction framework centered on low-dimensional and efficient offset tokens (LoDEOT), a five-dimensional offset token expression. As the forward process of offset token exhibit a natural denoising feature, we design a strategy to instill random noise to enhance the training.


On BONAI~\cite{wang_2023_learning}, LoDEOT achieves the highest roof-detection bAP and bAP50 among the compared end-to-end methods and leads all five offset-corrected footprint metrics, reaching 54.58\% FAP50 and a 5.23-pixel mEPE. Statistical analysis finds no significant performance differences in the tested width comparisons; we retain the geometry-motivated five-dimensional representation as a compact choice. Target-indexed denoising improves all reported ablation metrics, while the largest cumulative source set achieves the best five-domain macro-average results.

The contributions of this work are threefold:



\begin{itemize}
    \item Starting from first principles, we analyze the intrinsic relationship between the RFO and key physical parameters in the photogrammetric process, and tailor this process for off-nadir remote sensing scenario.
    \item Combining the photogrammetric analysis with domain insights into modern foundation models, we propose LoDEOT and equip it with a customized denoising training framework.
    \item Based on our design, we conduct extensive empirical evaluations to demonstrate the practical effectiveness of our method, and further corroborate our initial observations through the experimental phenomena.
\end{itemize}
\section{Related Work}

\paragraph{Raster and Polygonal Building Extraction}
Building extraction has progressed from semantic raster prediction on cross-city and large-scale benchmarks~\cite{maggiori_2017_semantic,demir_2018_deepglobe,ji_2019_fully} to explicit geometric outputs. Polygon-RNN++ recurrently predicts vertices and refines them with graph reasoning~\cite{acuna_2018_efficient}; joint semantic--geometric learning combines masks, corners, and edge orientations~\cite{li_2021_joint}. Frame Field Learning predicts contour directions for polygonization~\cite{girard_2021_polygonal}, whereas PolyWorld learns vertex connectivity through graph reasoning and differentiable matching~\cite{zorzi_2022_polyworld}. CVNet evolves building contours~\cite{xu_2022_cvnet}; BuildMapper learns initialization, deformation, and vertex reduction~\cite{wei_2023_buildmapper}; and HiSup couples vertex, boundary, and mask supervision~\cite{xu_2023_hisup}. PolyBuilding uses polygon queries and fixed-length point sequences~\cite{hu_2023_polybuilding}, while Pix2Poly generates polygon vertices and connectivity as a sequence~\cite{adimoolam_2025_pix2poly}. These methods improve boundary geometry but do not directly resolve the roof--footprint displacement specific to off-nadir imagery.

\paragraph{Off-Nadir Roof-to-Footprint Learning}
LOFT jointly predicts roof masks and RFOs and introduced BONAI~\cite{wang_2023_learning}; a multi-label ViT jointly models footprints, roofs, and visible building shapes~\cite{neupane_2025_multilabel}. OBM makes roof/offset extraction prompt driven~\cite{li_2024_promptdriven}, MLS-BRN uses multi-level supervision and a roof-offset-guided footprint extractor~\cite{li_2024_3d}, and PolyFootNet combines direct polygon prediction with mask prompting and self-offset attention~\cite{li_2025_polyfootnet}. Building on the promptable Segment Anything Model~\cite{kirillov_2023_segment}, RBF-SAM improves robustness to prompt type and quality~\cite{li_2026_rbfsam}. These advances change observations, outputs, or supervision. LoDEOT instead studies the capacity of the offset head itself. LOFT and MLS-BRN are the directly comparable roof-mask/RFO baselines under the reported BONAI protocol; prompt-dependent and polygon-output methods use different interfaces or evaluation settings.

\paragraph{Query-Based Representation and Denoising}
Instance segmentation progressed from Mask R-CNN's region-conditioned mask branch~\cite{he_2017_mask} to QueryInst's shared queries for categories, boxes, masks, and associations~\cite{fang_2021_instances}. DETR formulates detection as set prediction with object queries and bipartite matching~\cite{carion_2020_endtoend}; Conditional, Deformable, and DAB-DETR spatially condition them through conditional queries, sparse reference-point sampling, and dynamic anchors~\cite{meng_2021_conditional,zhu_2021_deformable,liu_2022_dabdetr}. MaskFormer uses mask classification, and Mask2Former restricts attention to predicted mask regions~\cite{cheng_2021_perpixel,cheng_2022_maskedattention}. DN-DETR reconstructs perturbed labels and boxes with known correspondence~\cite{li_2022_dndetr}; DINO adds contrastive denoising~\cite{zhang_2022_dino}, and Mask DINO unifies detection and segmentation with mask-enhanced denoising~\cite{li_2022_mask}. Existing denoising objectives focus on classes, boxes, and masks. LoDEOT updates a compact offset token at each layer by combining the previous offset token with current instance evidence, and uses known denoising target correspondences to supervise each layer's prediction against the same clean instance RFO.

\section{Method}
\label{sec:method}

\begin{figure*}[!t]
  \centering
  \includegraphics[width=\textwidth]{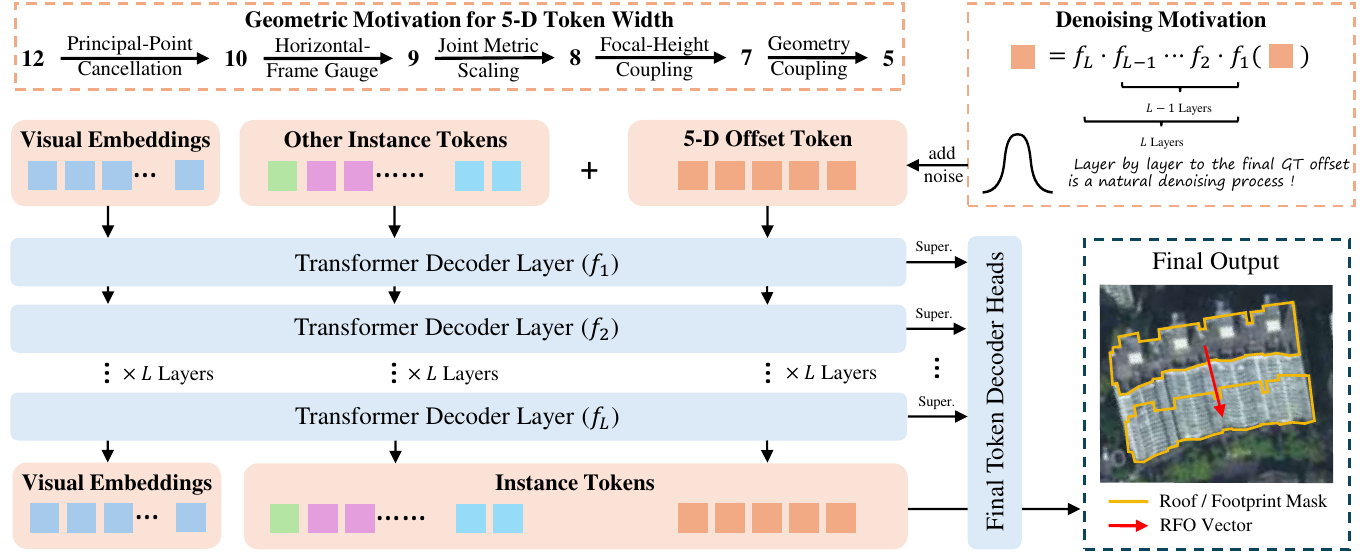}

  \caption{RFO prediction architecture after incorporating LoDEOT. Motivated by the projective geometry of 3D imaging, we redesign the offset-token representation. After the backbone encodes the visual features, each original high-dimensional instance token is augmented with a 5-D LoDEOT that carries the task-specific RFO representation, while retaining its roles in classification, roof-mask, and roof-box prediction. The augmented instance tokens are progressively updated through $L$ Transformer decoder layers; the final head predicts roof masks and RFOs to recover footprints. During training, each layer's RFO prediction is supervised by the same clean instance target.}
  \label{fig:architecture}

\end{figure*}

LoDEOT predicts roof instances and signed RFOs using high-dimensional instance tokens (decoder queries) and associated compact offset tokens (Figure~\ref{fig:architecture}).

\subsection{Overall Architecture}
\label{sec:problem}

Given an off-nadir image $\mathbf I$, an image encoder and an $L$-layer instance decoder produce $K$ query-indexed predictions:
\begin{equation}
  \widehat{\mathcal Y}
    =\left\{
      \left(
        \hat c_q,
        \hat{\mathbf b}_q,
        \hat{\mathbf m}_q,
        \hat{\boldsymbol\delta}_q
      \right)
    \right\}_{q=1}^{K}.
  \label{eq:problem_output}
\end{equation}
Here $\hat c_q$, $\hat{\mathbf b}_q$, and $\hat{\mathbf m}_q$ denote the class, roof box, and roof mask. The signed RFO $\boldsymbol\delta=\mathbf u_b-\mathbf u_r\in\mathbb R^2$ points from a roof reference point $\mathbf u_r$ to its corresponding footprint reference point $\mathbf u_b$. Both points and the displacement are expressed in the same pixel coordinate system.

Footprints are recovered by translating each predicted roof mask by $\hat{\boldsymbol\delta}_q$ under a local constant-displacement approximation. Multitask instance tokens retain their width; only offset-token width $d$ varies. Each offset token is updated from its previous value and current instance evidence.

\subsection{Geometry-Inspired Candidate Width}
\label{sec:dof_to_dim}

RFO depends on camera intrinsics, relative camera--building pose, and building height through shared combinations. We derive an exact five-parameter factorization of this map to guide offset-token design.

\textbf{Explicit Physical Offset Computation.}
Within a local instance neighborhood, let $\mathbf B\in\mathbb R^3$ be a footprint reference point, $\mathbf C\in\mathbb R^3$ the camera center, $\mathbf g\in\mathbb S^2$ the fixed world-vertical direction, and $H>0$ the building height. Vertical extrusion gives the roof point $\mathbf B+H\mathbf g$, while $\mathbf r=\mathbf B-\mathbf C$ removes the arbitrary world origin. Let $\mathbf R\in\mathrm{SO}(3)$ map world directions to the camera frame. A pinhole intrinsic matrix $\mathbf K$, normalized by $K_{33}=1$, contains focal scales $f_x,f_y>0$, principal point $(c_x,c_y)$, and image-axis skew $\kappa$. With $\pi([a,b,c]^\top)=[a/c,b/c]^\top$, the two projections and their signed difference are
\begin{equation}
  \begin{aligned}
  \mathbf u_b&=\pi(\mathbf K\mathbf R\mathbf r),\\
  \mathbf u_r&=\pi\!\left(\mathbf K\mathbf R(\mathbf r+H\mathbf g)\right),\\
  \boldsymbol\delta&=\mathbf u_b-\mathbf u_r
    =\Phi(\mathbf K,\mathbf R,\mathbf r,H).
  \end{aligned}
  \label{eq:offset_func}
\end{equation}
The local state has twelve coordinates: five intrinsics, three rotation coordinates, three components of $\mathbf r$, and $H$.

\emph{Five-parameter factorization of the physical map.}
\label{sec:five_factorization}
Define camera-frame vectors $\boldsymbol\xi=\mathbf R\mathbf r=(X,Y,Z)^\top$ and $\boldsymbol\nu=\mathbf R\mathbf g=(\nu_x,\nu_y,\nu_z)^\top\in\mathbb S^2$, normalized coordinates $x=X/Z$, $y=Y/Z$, and relative height $\rho=H/Z$. For both reference points to lie in front of the camera, require $Z>0$ and $Z+H\nu_z>0$. Subtracting their projections yields
\begin{align}
  \delta_x
    &=\frac{\rho}{1+\rho\nu_z}
      [f_x(x\nu_z-\nu_x)+\kappa(y\nu_z-\nu_y)],
      \label{eq:general_skew_offset_x}\\
  \delta_y
    &=\frac{f_y\rho}{1+\rho\nu_z}(y\nu_z-\nu_y).
      \label{eq:general_skew_offset_y}
\end{align}
Shared camera intrinsics make the principal point $(c_x,c_y)$ cancel from the projection difference, reducing the coordinate count from twelve to ten.

Pose enters only through $(\boldsymbol\xi,\boldsymbol\nu)$. For $\mathbf Q\in\mathrm{SO}(3)$ with $\mathbf Q\mathbf g=\mathbf g$, the transformation $(\mathbf R,\mathbf r)\mapsto(\mathbf R\mathbf Q^\top,\mathbf Q\mathbf r)$ preserves both vectors. This horizontal-frame gauge preserves physical viewing geometry and replaces six pose coordinates with three for $\boldsymbol\xi$ and two for $\boldsymbol\nu\in\mathbb S^2$, leaving nine overall.

Joint scaling $(\mathbf r,H)\mapsto(\lambda\mathbf r,\lambda H)$, $\lambda>0$, preserves projections; $(X,Y,Z,H)$ enters only through $(x,y,\rho)=(X/Z,Y/Z,H/Z)$, leaving eight coordinates: three intrinsics, $(x,y,\rho)$, and two for $\boldsymbol\nu$.

Define focal anisotropy $a=f_x/f_y$, normalized skew $\widetilde\kappa=\kappa/f_y$, and the composite amplitude $\gamma=f_y\rho/(1+\rho\nu_z)$ combining common focal scale and relative height. Then
\begin{equation}
  \boldsymbol\delta
    =\gamma
      \begin{bmatrix}
        a&\widetilde\kappa\\
        0&1
      \end{bmatrix}
      \begin{bmatrix}
        x\nu_z-\nu_x\\
        y\nu_z-\nu_y
      \end{bmatrix}.
  \label{eq:factorized_offset}
\end{equation}
To make this coupling explicit, let $\mathbf A=\bigl[\begin{smallmatrix}f_x&\kappa\\0&f_y\end{smallmatrix}\bigr]$. For any $\mu>0$ satisfying $\mu+(\mu-1)\rho\nu_z>0$, set $\mathbf A'=\mu\mathbf A$ and $\rho'=\rho/[\mu+(\mu-1)\rho\nu_z]$. This transformation preserves positive depth and satisfies
\begin{equation}
  \frac{\rho'}{1+\rho'\nu_z}\mathbf A'
    =\frac{\rho}{1+\rho\nu_z}\mathbf A.
  \label{eq:focal_height_invariance}
\end{equation}
Thus, at fixed $(x,y,\boldsymbol\nu)$, common focal scale and relative height affect the offset only through $\gamma$. The retained variables are $(a,\widetilde\kappa,\gamma,x,y,\boldsymbol\nu)$, with seven coordinates in total.

After introducing $\gamma$, the remaining positional and directional dependence enters through the geometric vector
\begin{equation}
  \boldsymbol\beta
    =
    \begin{bmatrix}
      \beta_x\\
      \beta_y
    \end{bmatrix}
    =
    \begin{bmatrix}
      x\nu_z-\nu_x\\
      y\nu_z-\nu_y
    \end{bmatrix}.
  \label{eq:geometric_factor}
\end{equation}
The four coordinates of $(x,y,\boldsymbol\nu)$ are therefore represented by two combinations while $\gamma$ is retained separately. For fixed $\boldsymbol\nu$ with $\nu_z\neq0$, the map from $(x,y)$ to $\boldsymbol\beta$ has Jacobian $\nu_z\mathbf I_2$, so these two geometry coordinates can vary independently. Combining the preceding steps gives
\begin{equation}
  \begin{aligned}
    12
      &\xrightarrow{-2\ \text{principal point}}
       10
       \xrightarrow{-1\ \text{horizontal gauge}}
       9,\\
    9
      &\xrightarrow{-1\ \text{metric scale}}
       8
       \xrightarrow{-1\ \text{focal--height}}
       7
       \xrightarrow{-2\ \text{geometry}}
       5,\\
    \boldsymbol\eta_5
      &=(a,\widetilde\kappa,\gamma,\beta_x,\beta_y)^\top,\\
    \boldsymbol\delta
      &=\Phi_5(\boldsymbol\eta_5)
       =\gamma
        \begin{bmatrix}
          a&\widetilde\kappa\\
          0&1
        \end{bmatrix}
        \boldsymbol\beta.
  \end{aligned}
  \label{eq:five_physical_descriptor}
\end{equation}
The counts describe retained factorization coordinates, not a minimum representation dimension. The description is exact under the pinhole-camera, vertical-extrusion, and positive-depth assumptions, including $\nu_z=0$, where $\boldsymbol\beta=-(\nu_x,\nu_y)^\top$ has unit norm and geometry is more restricted.

\textbf{From Physical Factorization to the Default Offset-Token Width.}
The five-parameter descriptor $\boldsymbol\eta_5$ organizes the general imaging map into intrinsic shape, displacement amplitude, and relative geometry, motivating $d=5$ as the default offset-token width. Learned channels need not correspond individually to physical parameters, and the readout need not implement $\Phi_5$. Under shared calibration and locally restricted viewing geometry, some factors can become shared or coupled, providing a possible explanation for similar performance at smaller widths. The supplementary material derives these conditional reductions, which inform the width analysis in Section~\ref{sec:width_ablation}.

\subsection{Mask-Guided Compact Offset Head}
\label{sec:offsethead}

We omit the query index and show layer subscripts only in recurrent updates; all vectors are columns. On a resized grid of at most $S\times S$ with $P$ locations, $\mathbf F,\mathbf E\in\mathbb R^{P\times d_m}$ denote mask features and sinusoidal positional encodings, and $\mathbf m\in\mathbb R^P$ roof-mask logits. With sigmoid $\sigma$ and stop-gradient operator $\operatorname{sg}$, set $\mathbf w=\sigma(\operatorname{sg}(\mathbf m))$ and pool features as
\begin{equation}
  \mathbf p=\frac{\mathbf w}{\mathbf 1^\top\mathbf w+\epsilon},
  \qquad
  \mathbf f=f_{\rm feat}\bigl((\mathbf F+\mathbf E)^\top\mathbf p\bigr).
  \label{eq:mask_pooled_feature}
\end{equation}
Here $\mathbf 1\in\mathbb R^P$ is the all-ones vector, $\epsilon>0$ stabilizes normalization, and the feature branch remains trainable. During training, selected queries receive zero-mean Gaussian noise on $\mathbf w$, clipped to $[0,1]$ before normalization.

Let $\mathbf X\in[0,1]^{P\times2}$ contain normalized grid coordinates, and let $\mathbf b\in\mathbb R^4$ be the detached normalized center--size roof box with center $\mathbf b_c\in\mathbb R^2$. The stabilized mask centroid is $\mathbf c=\mathbf X^\top\mathbf p$. Spatial evidence is constructed as
\begin{equation}
  s=\|\mathbf p\|_\infty
    \exp\!\left(\frac{\mathbf p^\top\log(\mathbf p+\epsilon\mathbf 1)}{P}\right),
  \qquad
  \boldsymbol\zeta=f_{\rm sp}\!\left(
    \operatorname{Concat}\bigl(\mathbf b,\mathbf c,s(\mathbf c-\mathbf b_c)\bigr)
  \right).
  \label{eq:spatial_cue}
\end{equation}
The logarithm acts elementwise, and $\operatorname{Concat}$ concatenates column vectors. The concentration score $s$ gates roof features and box--centroid displacement. It is sensitive to spatial support size and is not a calibrated mask-quality estimate.

At decoder layer $\ell$, OffsetHead updates the offset token using its previous value and current instance evidence:
\begin{equation}
  \mathbf o_\ell=f_{\rm tok}\!\left(
    \operatorname{Concat}\bigl(\mathbf o_{\ell-1},\mathbf h_\ell,s_\ell\mathbf f_\ell,\boldsymbol\zeta_\ell\bigr)
  \right),
  \qquad
  \hat{\mathbf z}_\ell=f_{\rm out}(\mathbf o_\ell).
  \label{eq:offset_head_layers}
\end{equation}
Here $\mathbf h_\ell\in\mathbb R^D$ is the instance token, $\mathbf o_\ell\in\mathbb R^d$ the offset token, and $\hat{\mathbf z}_\ell\in\mathbb R^2$ the normalized RFO prediction defined below. The maps $f_{\rm feat}$, $f_{\rm sp}$, and $f_{\rm tok}$ are learned, and $f_{\rm out}:\mathbb R^d\to\mathbb R^2$ is a linear readout. The offset token passes to the next decoder layer, with $d=5$ as the default width.

\subsection{Target-Indexed Layer-Wise RFO Supervision}
\label{sec:offset_supervision}

Under invertible affine augmentation $T(\mathbf u)=\mathbf J\mathbf u+\mathbf t$, encode the pixel-space RFO as $\mathbf z=\alpha\mathbf D_a^{-1}\mathbf J\boldsymbol\delta$, where $\mathbf D_a=\operatorname{diag}(W_a,H_a)$ contains augmented image dimensions and $\alpha>0$. For mask translation, decode $\alpha^{-1}\mathbf D_a\hat{\mathbf z}$ in augmented coordinates, applying $\mathbf J^{-1}$ when returning to original coordinates.

Denoising groups replicate ground-truth instances with perturbed labels and boxes. Attention masks isolate groups and block matching queries' access to denoising queries. Known target indices supervise successive RFOs against the same clean target without additional matching. For valid $\mathbf z_n$, a fixed lift $\mathbf U\in\mathbb R^{d\times2}$ ($d\ge2$, $\mathbf U^\top\mathbf U=\mathbf I_2$) initializes offset tokens as $\mathbf U\mathbf z_n+\boldsymbol\varepsilon_q$ (additive) or $(\mathbf U\mathbf z_n)\odot(\mathbf 1_d+\boldsymbol\varepsilon_q)$ (multiplicative). Here $\boldsymbol\varepsilon_q\in\mathbb R^d$ is the sampled perturbation, $\mathbf 1_d$ the all-ones vector, and $\odot$ elementwise multiplication. Matching-query offset tokens are initialized from instance features; both query types follow Eq.~\eqref{eq:offset_head_layers}.

For supervision, write $\hat{\mathbf z}_\ell(q)$ for the normalized RFO predicted by query $q$ at layer $\ell$, and $\mathbf z_n$ for target $n$. The coordinate-summed regression loss is
\begin{equation}
  \ell_\delta(\hat{\mathbf z},\mathbf z)
  =\sum_{k=1}^{2}\operatorname{SmoothL1}(\hat z_k-z_k),
  \qquad
  \operatorname{SmoothL1}(t)=
  \begin{cases}
    \frac{1}{2}t^2, & |t|<1,\\
    |t|-\frac{1}{2}, & |t|\ge1.
  \end{cases}
  \label{eq:offset_penalty}
\end{equation}

Let $\mathcal M_\ell$ contain the Hungarian-matched query--target pairs at layer $\ell$, and let $\mathcal D$ contain the fixed denoising pairs across all groups. With RFO-validity indicator $v_n\in\{0,1\}$, the matching-query and denoising RFO losses are
\begin{equation}
  \begin{aligned}
    \mathcal L_\delta(\ell)
      &=\frac{1}{N}\sum_{(q,n)\in\mathcal M_\ell}
        v_n\,\ell_\delta\bigl(\hat{\mathbf z}_\ell(q),\mathbf z_n\bigr),\\
    \mathcal L_{\rm DN}
      &=\frac{1}{NG}\sum_{\ell\in\mathcal S}\omega_\ell
        \sum_{(q,n)\in\mathcal D}
        v_n\,\ell_\delta\bigl(\hat{\mathbf z}_\ell(q),\mathbf z_n\bigr).
  \end{aligned}
  \label{eq:offset_objectives}
\end{equation}
Here $N$ is the ground-truth normalization count clamped to at least one, $G$ the denoising-group count, and $\mathcal S$ the supervised denoising stages. Each $\omega_\ell$ includes both stage and RFO-loss weights. Invalid RFOs ($v_n=0$) use the placeholder $\mathbf z_n=\mathbf0$ and contribute neither regression nor matching cost, without changing $N$. $\mathcal L_{\rm DN}$ includes only RFO supervision.

Hungarian matching combines classification, point-sampled mask BCE/Dice, box $L_1$, and GIoU costs with the valid-only RFO cost $v_n\|\hat{\mathbf z}_\ell(q)-\mathbf z_n\|_1$, weighted by $\lambda_{\rm box}^{m}\lambda_\delta^{m}$. The joint set-prediction loss at each supervised stage is
\begin{equation}
  \begin{aligned}
    \mathcal L_{\rm set}(\ell)
      ={}&
      \lambda_{\rm cls}\mathcal L_{\rm cls}(\ell)
      +\lambda_{\rm mask}\mathcal L_{\rm mask}(\ell)
      +\lambda_{\rm dice}\mathcal L_{\rm dice}(\ell)\\
      &+\lambda_{\rm box}\mathcal L_{\rm box}(\ell)
      +\lambda_{\rm giou}\mathcal L_{\rm giou}(\ell)
      +\lambda_\delta\mathcal L_\delta(\ell).
  \end{aligned}
  \label{eq:set_loss}
\end{equation}
The terms correspond to classification, mask BCE, mask Dice, box $L_1$, GIoU, and RFO regression, respectively. Final, auxiliary, and encoder outputs receive the corresponding set losses; the denoising branch additionally receives the RFO loss $\mathcal L_{\rm DN}$.
\section{Experiments}
\label{sec:experiments}

\subsection{Experimental Settings}
\label{sec:experimental_settings}

\paragraph{Datasets}
We use five named data sources. BONAI~\cite{wang_2023_learning} supports the main comparison, IRSA~\cite{meng_2025_irsamap} the ablations, and BONAI, IRSA, WHU~\cite{ji_2019_fully}, and UAV~\cite{li_2026_oblicity} the cumulative-source study; Huizhou~\cite{li_2024_promptdriven} is held out from training for cross-domain evaluation.

\paragraph{Implementation Details}
Unless stated otherwise, single-source models use an ImageNet-22K-pretrained Swin-Base backbone~\cite{liu_2021_swin}, nine decoder layers, 300 queries, and $d=5$. We optimize with AdamW~\cite{loshchilov_2019_decoupled} for 35 epochs on six NVIDIA RTX 3090 GPUs. Inputs use a $768\times768$ canvas with scale jittering, flipping, and rotation. All ablations use the same IRSA split and training protocol.

\paragraph{Evaluation Metrics}
Following the roof-mask/RFO protocol~\cite{wang_2023_learning,li_2024_3d}, bAP/sAP evaluate roof detection/instance segmentation and bAP50/sAP50 use IoU $0.5$. FAP50 and best-threshold F1 evaluate offset-corrected footprints, while Boundary IoU (B-IoU) measures boundary consistency. Mean endpoint error (mEPE) is the average pixel-space RFO endpoint error; PCP@5 is the percentage below five pixels. Higher is better except for mEPE.

\subsection{Comparison with Existing Methods}
\label{sec:quantitative_comparison}

\begin{figure}[!t]
    \centering
    \includegraphics[width=\columnwidth]{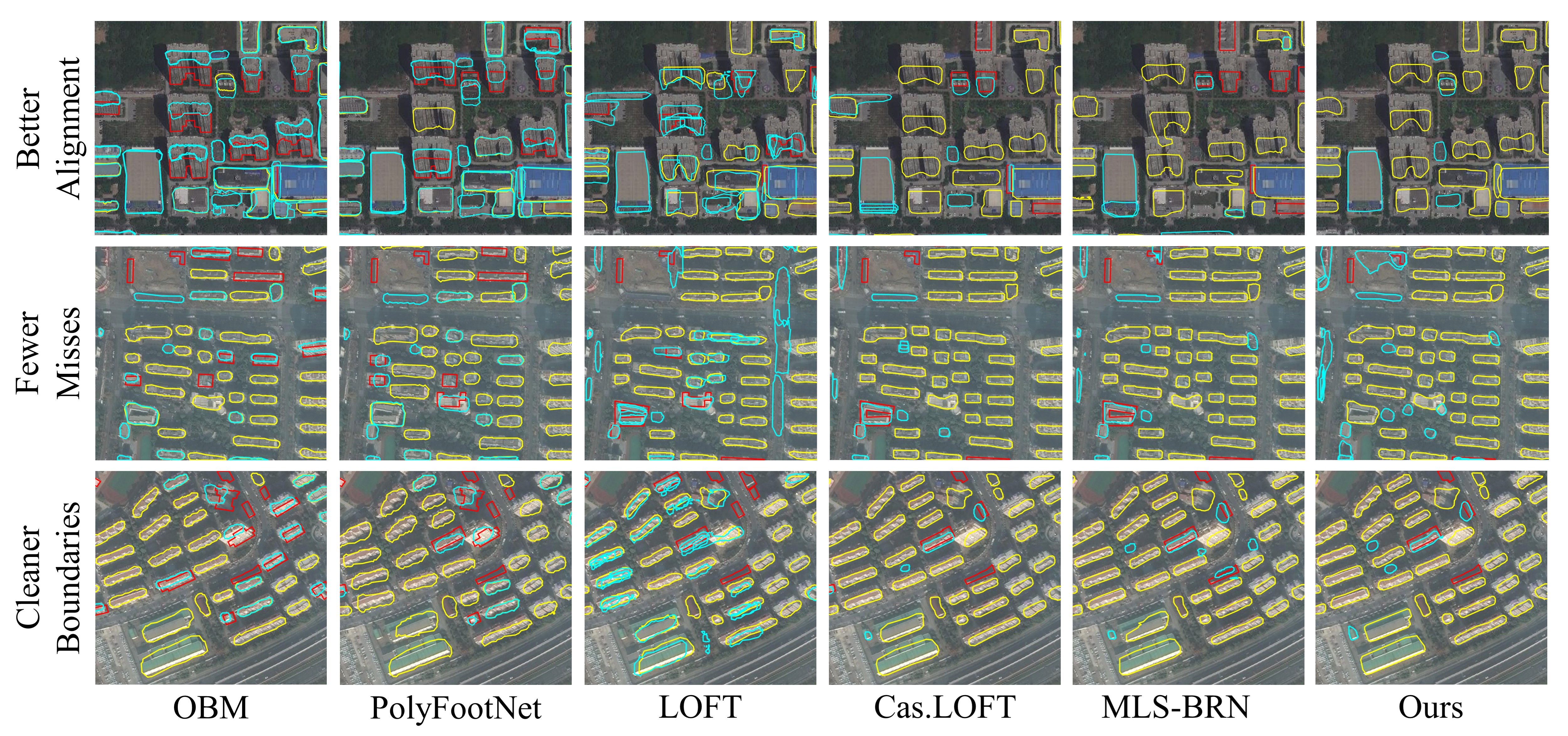}

    \caption{Qualitative comparison of OBM, PolyFootNet, LOFT, Cas.LOFT, MLS-BRN, and LoDEOT (ours) on three representative BONAI scenes. Each row shows the same scene across methods, and each column corresponds to one method. Yellow denotes true-positive matches (IoU $\geq 0.5$), cyan unmatched predictions (false positives), and red unmatched ground-truth instances (false negatives). OBM and PolyFootNet use LoDEOT-predicted masks as guidance.}
    \label{fig:qualitative_comparison}

\end{figure}

Table~\ref{tab:bonai_comparison} compares LoDEOT with end-to-end
LOFT~\cite{wang_2023_learning}, Cas.LOFT~\cite{li_2024_promptdriven},
and MLS-BRN~\cite{li_2024_3d} on BONAI.
OBM~\cite{li_2024_promptdriven} and
PolyFootNet~\cite{li_2025_polyfootnet} use LoDEOT-predicted masks
as instance guidance and serve as supplementary references.
The end-to-end baselines share roof-mask and signed-RFO outputs,
but architecture and training differences make this a system-level
comparison rather than an isolated test of the compact offset head.
Figure~\ref{fig:qualitative_comparison} shows all six methods
on the same three scenes.

\begin{table*}[tp]
\caption{Performance comparison of externally guided and end-to-end methods on the BONAI test set. The best results among the end-to-end methods are highlighted. All metrics are percentages except mEPE, which is measured in pixels.}
\label{tab:bonai_comparison}
\begin{center}
{\normalsize
\setlength{\tabcolsep}{2.3pt}
\begin{tabular}{@{}lccccccccc@{}}
\toprule
\multirow{2}{*}{Method}
& \multicolumn{4}{c}{Roof Detection and Segmentation}
& \multicolumn{5}{c}{Offset-Corrected Footprint} \\
\cmidrule(lr){2-5}
\cmidrule(lr){6-10}
& bAP$\uparrow$
& bAP50$\uparrow$
& sAP$\uparrow$
& sAP50$\uparrow$
& FAP50$\uparrow$
& F1$\uparrow$
& B-IoU$\uparrow$
& mEPE$\downarrow$
& PCP@5$\uparrow$ \\
\midrule
\multicolumn{10}{>{\columncolor[gray]{0.92}}l}
{\color{black}\itshape Externally Guided Methods} \\

\graytxt{OBM}
& \graytxt{--} & \graytxt{--}
& \graytxt{12.59} & \graytxt{36.71}
& \graytxt{22.21} & \graytxt{41.72}
& \graytxt{9.25} & \graytxt{5.72} & \graytxt{53.68} \\

\graytxt{PolyFootNet}
& \graytxt{--} & \graytxt{--}
& \graytxt{8.91} & \graytxt{33.18}
& \graytxt{18.76} & \graytxt{39.75}
& \graytxt{7.32} & \graytxt{6.54} & \graytxt{44.71} \\

\midrule
\multicolumn{10}{>{\columncolor[gray]{0.92}}l}
{\itshape End-to-End Methods} \\

LOFT
& 10.50 & 42.90 & 21.00 & 59.50
& 37.73 & 49.64 & 8.78 & 5.65 & 57.74 \\

Cas.LOFT
& 15.80 & 51.45 & \textbf{30.21} & \textbf{63.25}
& 47.02 & 55.60 & 11.12 & 5.52 & 60.76 \\

MLS-BRN
& 15.45 & 51.63 & 30.20 & 62.81
& 42.73 & 53.99 & 11.17 & 5.67 & 61.02 \\

LoDEOT(Ours)
& \textbf{31.38} & \textbf{57.20} & 26.37 & 54.94
& \textbf{54.58} & \textbf{58.87} & \textbf{11.20}
& \textbf{5.23} & \textbf{64.25} \\
\bottomrule
\end{tabular}
}
\end{center}
\end{table*}

LoDEOT leads the end-to-end methods in bAP/bAP50 (31.38/57.20) and all five footprint metrics, and outperforms both externally guided references on all reported metrics. However, its sAP/sAP50 trail the roof-segmentation leader Cas.LOFT by 3.84/8.31 points, indicating stronger roof localization and RFO recovery without uniformly better roof masks (Table~\ref{tab:bonai_comparison}).

\subsection{Cumulative Multi-Source Training and Cross-Domain Evaluation}
\label{sec:multi_source}

We construct cumulative levels BONAI (L1), BONAI+\allowbreak IRSA (L2), BONAI+\allowbreak IRSA+\allowbreak WHU (L3), and BONAI+\allowbreak IRSA+\allowbreak WHU+\allowbreak UAV (L4), and evaluate each on all four sources plus unseen Huizhou. Table~\ref{tab:cumulative_sources} reports sAP50/FAP50 and their unweighted five-domain averages.

\begin{table*}[tp]
\caption{Evaluation of LoDEOT with $d=5$ under cumulative multi-source training.  The average is the unweighted mean over the five test domains.}
\label{tab:cumulative_sources}
\begin{center}
{\normalsize
\setlength{\tabcolsep}{1.5pt}
\begin{tabular}{@{}c*{12}{r}@{}}
\toprule
\multirow{2}{*}{Level}
& \multicolumn{2}{c}{BONAI}
& \multicolumn{2}{c}{IRSA}
& \multicolumn{2}{c}{WHU}
& \multicolumn{2}{c}{UAV}
& \multicolumn{2}{c}{Huizhou}
& \multicolumn{2}{c}{Average} \\
\cmidrule(lr){2-3}
\cmidrule(lr){4-5}
\cmidrule(lr){6-7}
\cmidrule(lr){8-9}
\cmidrule(lr){10-11}
\cmidrule(lr){12-13}
& sAP50 & FAP50
& sAP50 & FAP50
& sAP50 & FAP50
& sAP50 & FAP50
& sAP50 & FAP50
& sAP50 & FAP50 \\
\midrule
L1
& 54.94 & \textbf{54.58} & 9.70 & 8.77 & 49.50 & 48.75
& 31.64 & 28.49 & 35.72 & 31.17 & 36.30 & 34.35 \\
L2
& 54.68 & 51.58 & \textbf{51.67} & \textbf{48.76} & 26.65 & 26.60
& 14.03 & 12.98 & 8.82 & 5.56 & 31.17 & 29.09 \\
L3
& 54.21 & 51.76 & 49.42 & 46.30 & \textbf{96.08} & \textbf{96.08}
& 24.25 & 22.65 & 31.22 & 25.01 & 51.04 & 48.36 \\
L4
& \textbf{55.22} & 52.29 & 49.27 & 46.34 & 95.55 & 95.54
& \textbf{44.99} & \textbf{41.75} & \textbf{39.13} & \textbf{32.28} & \textbf{56.83} & \textbf{53.64} \\
\bottomrule
\end{tabular}
}
\end{center}
\end{table*}

Each added source gives its largest immediate FAP50 gain on its own domain. Performance is non-monotonic: IRSA and WHU peak at L2 and L3; BONAI peaks at L4 for sAP50 and L1 for FAP50, while UAV and unseen Huizhou peak at L4 for both. Relative to L1, L4 improves the macro-average sAP50/FAP50 by 20.53/19.29 points to 56.83/53.64 and Huizhou scores by 3.41/1.11 points. Since source composition, sample count, and training exposure vary jointly, these results describe cumulative-source effects rather than a controlled scaling law.

\subsection{Ablation Studies}
\label{sec:ablation}

\paragraph{Offset-Token Width}
\label{sec:width_ablation}

Table~\ref{tab:width_ablation} evaluates 18 widths with five seeds each. $d=7$ has the highest means, but increasing width does not monotonically improve performance. Paired two-sided $t$-tests comparing $d=5$ with other widths detect no significant differences on the six metrics (raw $p\geq0.098$; Holm correction over 102 tests). This does not establish equivalence or optimality.

The proprietary IRSA collection uses a fixed satellite-imaging modality. Under shared calibration and processing with locally restricted geometry, some physical factors can become shared or coupled; Appendix~\ref{app:restricted_geometry} derives these conditional reductions and their error bounds. This offers a possible explanation for similar results at $d=2$--$4$, while $d=5$ remains the default motivated by the general five-parameter organization.

\begin{table}[tp]
\caption{Offset-token width ablation on the IRSA test set. Results are mean $\pm$ sample standard deviation over five random seeds.}
\label{tab:width_ablation}
\begin{center}
{\normalsize
\setlength{\tabcolsep}{2pt}
\begin{tabular}{@{}c*{6}{c}@{}}
\toprule
$d$ & bAP$\uparrow$ & bAP50$\uparrow$ & sAP$\uparrow$ & sAP50$\uparrow$ & FAP50$\uparrow$ & F1$\uparrow$ \\
\midrule
2 & $29.66\pm1.20$ & $49.10\pm1.56$ & $28.66\pm1.20$ & $48.96\pm1.67$ & $45.67\pm1.60$ & $51.60\pm1.20$ \\
3 & $29.63\pm0.82$ & $48.96\pm1.42$ & $28.73\pm0.83$ & $49.01\pm1.48$ & $45.76\pm1.28$ & $51.62\pm1.34$ \\
4 & $29.04\pm1.25$ & $48.03\pm2.04$ & $28.16\pm1.19$ & $48.03\pm2.06$ & $45.07\pm1.94$ & $50.97\pm1.55$ \\
5 & $29.42\pm1.50$ & $48.54\pm2.35$ & $28.41\pm1.47$ & $48.38\pm2.46$ & $45.20\pm2.52$ & $50.93\pm2.16$ \\
6 & $30.08\pm0.97$ & $49.68\pm0.74$ & $29.10\pm0.76$ & $49.62\pm0.72$ & $46.32\pm0.98$ & $51.97\pm0.69$ \\
7 & $\mathbf{30.61}\pm0.51$ & $\mathbf{50.31}\pm0.79$ & $\mathbf{29.62}\pm0.48$ & $\mathbf{50.35}\pm0.89$ & $\mathbf{47.08}\pm0.76$ & $\mathbf{52.91}\pm0.66$ \\
8 & $28.74\pm2.28$ & $47.81\pm3.43$ & $27.80\pm2.30$ & $47.86\pm3.28$ & $44.46\pm3.30$ & $50.62\pm2.33$ \\
9 & $29.60\pm1.10$ & $49.07\pm2.00$ & $28.62\pm1.22$ & $49.03\pm2.13$ & $45.87\pm2.19$ & $51.66\pm1.74$ \\
10 & $29.66\pm0.94$ & $49.25\pm1.47$ & $28.71\pm0.86$ & $49.16\pm1.43$ & $46.02\pm1.33$ & $51.70\pm1.22$ \\
11 & $29.58\pm0.97$ & $48.99\pm1.36$ & $28.69\pm0.92$ & $48.93\pm1.33$ & $45.86\pm1.34$ & $51.37\pm1.11$ \\
12 & $29.91\pm0.51$ & $49.39\pm0.39$ & $29.05\pm0.33$ & $49.42\pm0.31$ & $46.33\pm0.36$ & $51.99\pm0.16$ \\
13 & $29.89\pm1.22$ & $49.41\pm1.84$ & $28.88\pm1.26$ & $49.32\pm1.96$ & $46.20\pm2.06$ & $52.02\pm1.96$ \\
14 & $29.56\pm1.09$ & $48.77\pm1.65$ & $28.55\pm1.00$ & $48.75\pm1.65$ & $45.58\pm1.58$ & $51.39\pm1.41$ \\
15 & $29.61\pm1.37$ & $48.92\pm2.31$ & $28.64\pm1.47$ & $48.88\pm2.29$ & $45.71\pm2.04$ & $51.25\pm1.80$ \\
16 & $29.79\pm0.45$ & $49.19\pm0.79$ & $28.94\pm0.32$ & $49.29\pm0.80$ & $46.02\pm0.67$ & $51.74\pm0.67$ \\
32 & $29.68\pm0.86$ & $49.22\pm1.28$ & $28.87\pm0.79$ & $49.32\pm1.27$ & $46.07\pm1.28$ & $51.56\pm1.21$ \\
64 & $30.12\pm0.85$ & $49.54\pm1.63$ & $29.15\pm0.96$ & $49.55\pm1.61$ & $46.66\pm1.39$ & $52.23\pm1.32$ \\
256 & $29.92\pm1.05$ & $49.47\pm1.64$ & $28.95\pm1.07$ & $49.47\pm1.64$ & $46.26\pm1.63$ & $51.72\pm1.24$ \\
\bottomrule
\end{tabular}
}
\end{center}
\end{table}

\paragraph{Representation and Readout Cost}
Relative to $d=256$, $d=5$ reduces offset-token coordinates by $98.05\%$ and parameters per distinct affine readout ($2d+2$) from 514 to 12, a $97.67\%$ reduction (Appendix~\ref{app:parameter_accounting}), with mean FAP50/F1 lower by 1.06/0.79 points (Table~\ref{tab:width_ablation}).

\paragraph{Layer-Wise RFO Denoising}
\label{sec:offset_supervision_ablation}

Both denoising modes improve all nine metrics over the disabled path (Table~\ref{tab:offset_supervision_ablation}). Additive perturbation raises FAP50/F1 by 9.58/8.63 points and reduces mEPE from 3.29 to 2.71 pixels; relative to multiplicative perturbation, it improves FAP50/F1 by 2.58/2.30 points and reduces mEPE by 0.33 pixels. Additive leads eight metrics, with multiplicative ahead by 0.02 points in B-IoU, so additive is the default. The enabled--disabled comparison evaluates perturbation and supervision jointly, without isolating their effects.

\begin{table}[tp]
\caption{Ablation of the target-indexed RFO denoising path on the IRSA test set with $d=5$. Multiplicative and additive denote perturbations applied to the initial offset token. }
\label{tab:offset_supervision_ablation}
\begin{center}
{\normalsize
\setlength{\tabcolsep}{2.3pt}
\begin{tabular}{@{}lccccccccc@{}}
\toprule
Configuration
& bAP$\uparrow$
& bAP50$\uparrow$
& sAP$\uparrow$
& sAP50$\uparrow$
& FAP50$\uparrow$
& F1$\uparrow$
& B-IoU$\uparrow$
& mEPE$\downarrow$
& PCP@5$\uparrow$ \\
\midrule
Disabled
& 24.02 & 40.59 & 23.06 & 40.57
& 37.21 & 43.91 & 21.13 & 3.29 & 84.58 \\
Multiplicative
& 28.33 & 47.32 & 27.49 & 47.66
& 44.21 & 50.24 & \textbf{22.29} & 3.04 & 85.52 \\
Additive
& \textbf{30.27} & \textbf{49.57} & \textbf{29.17} & \textbf{49.65}
& \textbf{46.79} & \textbf{52.54} & 22.27
& \textbf{2.71} & \textbf{88.09} \\
\bottomrule
\end{tabular}
}
\end{center}
\end{table}

\subsection{Discussion}
\label{sec:discussion}

\paragraph{Experimental Evidence and Operation}
The comparisons evaluate complete systems and the joint denoising path; mask pooling, concentration gating, and box--mask geometry are not independently validated. Inference updates compact offset tokens from image, query, roof-mask, and box evidence; denoising groups, target indices, and lifted noisy states are training-only. Each predicted RFO applies a constant roof-mask translation without camera recovery. Whole-model cost also includes the shared encoder and decoder.

\paragraph{Dimensionality and Scope}
The five-factor organization motivates $d=5$ without fixed physical meanings for learned channels; restricted acquisition conditions can explain smaller widths' effectiveness. The factorization assumes local pinhole projection and vertical extrusion. Within-tile displacement directions can vary in UAV imagery~\cite{li_2026_oblicity}, limiting a shared coefficient and motivating broader width comparisons. Uncertain roof masks can weaken pooled appearance and centroid cues, so better footprints need not imply better roof segmentation.

\paragraph{Broader Implications}
LoDEOT addresses geometric alignment in building instance segmentation, separating appearance and displacement representations to connect roof delineation with footprint localization through a compact design.
\section{Conclusion}
\label{sec:conclusion}

We derive an exact five-parameter RFO factorization under local pinhole projection and vertical extrusion, providing a photogrammetric basis for compact offset-token design. Guided by this analysis, LoDEOT combines a mask-guided five-dimensional offset token with target-indexed layer-wise denoising while preserving high-dimensional multitask instance tokens. LoDEOT leads the compared end-to-end methods on all five BONAI footprint metrics; denoising improves all ablation metrics, and the largest cumulative source mixture achieves the best five-domain macro-average. Width comparisons detect no significant differences from $d=5$; conditional factor sharing and coupling may explain similar performance at smaller widths. Together, these findings support separating appearance modeling and geometric displacement prediction through compact offset representations and readouts for building instance segmentation.

\FloatBarrier

\subsection*{AI use statement}
In preparing this manuscript, we used generative AI tools to assist with English writing and translation, check technical terminology, and improve clarity and readability. The authors reviewed and revised the AI-assisted text and take responsibility for the final manuscript.

\bibliographystyle{IEEEtran}
\bibliography{ref1}

\clearpage
\appendices
\numberwithin{equation}{section}
\numberwithin{table}{section}
\numberwithin{figure}{section}
\section{Conditional Reduction of the RFO Map under Restricted Imaging Geometry}
\label{app:restricted_geometry}

This appendix derives the local approximation $\boldsymbol\delta\simeq H\mathbf k_0$ to help interpret the width study in Section~\ref{sec:width_ablation} and bounds its approximation error. The analysis uses the local pinhole and vertical-extrusion assumptions of Section~\ref{sec:dof_to_dim}. It distinguishes shared camera parameters, weak height-dependent perspective effects, and locally restricted geometry; these conditions need not hold uniformly across a dataset. Vector norms are Euclidean and matrix norms are the induced spectral norms.

\subsection{Exact Offset and Shared Calibration}
\label{app:exact_local_offset}

Let $(X,Y,Z)^\top$ be the camera-frame footprint point and $\boldsymbol\nu=(\nu_x,\nu_y,\nu_z)^\top\in\mathbb S^2$ the camera-frame vertical direction. A roof reference point has camera coordinates $(X,Y,Z)^\top+H\boldsymbol\nu$, where $H>0$. Require $Z>0$ and $Z+H\nu_z>0$. Write $\mathbf x=(X/Z,Y/Z)^\top$, $\boldsymbol\nu_{xy}=(\nu_x,\nu_y)^\top$, and
\begin{equation}
  \mathbf A=
  \begin{bmatrix}
    f_x&\kappa\\
    0&f_y
  \end{bmatrix}
  =f_y
  \begin{bmatrix}
    a&\widetilde\kappa\\
    0&1
  \end{bmatrix},
  \qquad
  \boldsymbol\beta=\nu_z\mathbf x-\boldsymbol\nu_{xy}.
  \label{eq:app_camera_geometry}
\end{equation}
Here $f_x,f_y>0$, $a=f_x/f_y$, and $\widetilde\kappa=\kappa/f_y$, consistently with Section~\ref{sec:dof_to_dim}. The principal point cancels when subtracting the two projections, giving
\begin{equation}
  \begin{aligned}
    \boldsymbol\delta
      &=\mathbf A\left(
        \mathbf x-
        \frac{Z\mathbf x+H\boldsymbol\nu_{xy}}{Z+H\nu_z}
        \right)\\
      &=\frac{H}{Z+H\nu_z}\mathbf A\boldsymbol\beta
       =\frac{H/Z}{1+(H/Z)\nu_z}\mathbf A\boldsymbol\beta.
  \end{aligned}
  \label{eq:app_exact_offset}
\end{equation}
This is the signed roof-to-footprint displacement in Eq.~\eqref{eq:factorized_offset}. Under a shared effective calibration and processing regime, $a$ and $\widetilde\kappa$ are shared across instances. For the local bound below, we hold the full matrix $\mathbf A$ fixed within a neighborhood; different neighborhoods or acquisitions may have different matrices. A fixed imaging modality or a standardized ground sampling distance alone does not establish this calibration condition.

\subsection{Height-Dependent Perspective Remainder}
\label{app:perspective_bound}

Define $\rho=H/Z$, $\tau=\rho\nu_z$, and the displacement obtained by omitting the height-dependent denominator, $\boldsymbol\delta_{\rm lin}=\rho\mathbf A\boldsymbol\beta$. Equation~\eqref{eq:app_exact_offset} gives the exact identity
\begin{equation}
  \begin{aligned}
    \boldsymbol\delta-\boldsymbol\delta_{\rm lin}
      &=-\tau\boldsymbol\delta,\\
    \frac{\|\boldsymbol\delta-\boldsymbol\delta_{\rm lin}\|}
         {\|\boldsymbol\delta\|}
      &=|\tau|\leq\rho
      \qquad(\boldsymbol\delta\neq\mathbf0).
  \end{aligned}
  \label{eq:app_relative_perspective_error}
\end{equation}
Thus, small $H/Z$ controls this perspective correction, independently of the focal scale. For a uniform bound, suppose $H\leq H_{\max}$ and $Z\geq Z_{\min}>0$, and set $\eta=H_{\max}/Z_{\min}<1$. Then $|\tau|\leq\eta$ and $1+\tau\geq1-\eta$, so
\begin{equation}
  \|\boldsymbol\delta-\boldsymbol\delta_{\rm lin}\|
    \leq\frac{\eta}{1-\eta}\|\boldsymbol\delta_{\rm lin}\|.
  \label{eq:app_absolute_perspective_error}
\end{equation}
This absolute bound also covers zero displacement. It controls the approximation within the assumed local camera model, not errors from replacing a physical imaging system by that model.

\subsection{Local Variation of the Geometric Factor}
\label{app:local_geometry_bound}

Choose a reference state $(Z_0,\mathbf x_0,\boldsymbol\nu_0)$ with $Z_0>0$, and define $\boldsymbol\beta_0=\nu_{0z}\mathbf x_0-\boldsymbol\nu_{0,xy}$. The reference subscripts denote components of $\boldsymbol\nu_0$. Direct subtraction gives
\begin{equation}
  \begin{aligned}
    \boldsymbol\beta-\boldsymbol\beta_0
      ={}&\nu_z(\mathbf x-\mathbf x_0)
       +(\nu_z-\nu_{0z})\mathbf x_0
       -(\boldsymbol\nu_{xy}-\boldsymbol\nu_{0,xy}),\\
    \|\boldsymbol\beta-\boldsymbol\beta_0\|
      \leq{}&\|\mathbf x-\mathbf x_0\|
        +(\|\mathbf x_0\|+1)
          \|\boldsymbol\nu-\boldsymbol\nu_0\|.
  \end{aligned}
  \label{eq:app_geometry_variation}
\end{equation}
The inequality follows from $|\nu_z|\leq1$ and the triangle inequality. If $\|\mathbf x-\mathbf x_0\|\leq\varepsilon_x$ and $\|\boldsymbol\nu-\boldsymbol\nu_0\|\leq\varepsilon_\nu$, then $\|\boldsymbol\beta-\boldsymbol\beta_0\|\leq\varepsilon_\beta$, where $\varepsilon_\beta=\varepsilon_x+(\|\mathbf x_0\|+1)\varepsilon_\nu$. For exactly shared viewing orientation, the difference reduces to $\nu_z(\mathbf x-\mathbf x_0)$; small normalized spatial extent therefore limits geometric variation even when building heights differ.

\subsection{A Shared Local Coefficient and Its Error Bound}
\label{app:shared_coefficient}

Assume in addition that $|Z-Z_0|\leq\varepsilon_Z Z_0$ with $0\leq\varepsilon_Z<1$. Define the locally shared coefficient vector $\mathbf k_0=\mathbf A\boldsymbol\beta_0/Z_0$. It has units of image displacement per unit building height. Adding and subtracting $\mathbf A\boldsymbol\beta_0/Z$ gives
\begin{equation}
  \begin{aligned}
    \frac{\mathbf A\boldsymbol\beta}{Z}-\mathbf k_0
      &=\frac{\mathbf A(\boldsymbol\beta-\boldsymbol\beta_0)}{Z}
        +\mathbf A\boldsymbol\beta_0
          \left(\frac{1}{Z}-\frac{1}{Z_0}\right),\\
    \left\|\frac{\mathbf A\boldsymbol\beta}{Z}-\mathbf k_0\right\|
      &\leq\frac{\|\mathbf A\|}{Z_0(1-\varepsilon_Z)}
        \left(
          \varepsilon_\beta+\varepsilon_Z\|\boldsymbol\beta_0\|
        \right).
  \end{aligned}
  \label{eq:app_coefficient_error}
\end{equation}
Indeed, $Z\geq Z_0(1-\varepsilon_Z)$ and $|1/Z-1/Z_0|\leq\varepsilon_Z/[Z_0(1-\varepsilon_Z)]$. Since $\boldsymbol\delta_{\rm lin}=H\mathbf A\boldsymbol\beta/Z$, the residual $\mathbf e=\boldsymbol\delta-H\mathbf k_0$ is the sum of $\boldsymbol\delta-\boldsymbol\delta_{\rm lin}$ and $H(\mathbf A\boldsymbol\beta/Z-\mathbf k_0)$. Combining Eqs.~\eqref{eq:app_absolute_perspective_error} and~\eqref{eq:app_coefficient_error}, with $\|\boldsymbol\beta\|\leq\|\boldsymbol\beta_0\|+\varepsilon_\beta$, yields
\begin{equation}
  \begin{aligned}
    \boldsymbol\delta&=H\mathbf k_0+\mathbf e,\\
    \|\mathbf e\|
      &\leq\frac{H\|\mathbf A\|}{Z_0(1-\varepsilon_Z)}
       \left[
         \frac{\eta}{1-\eta}
           \left(\|\boldsymbol\beta_0\|+\varepsilon_\beta\right)
         +\varepsilon_\beta
         +\varepsilon_Z\|\boldsymbol\beta_0\|
       \right].
  \end{aligned}
  \label{eq:app_shared_coefficient_bound}
\end{equation}
The three terms account for height-dependent perspective, local geometric variation, and local depth variation. Their smallness gives a controlled absolute approximation; when the leading displacement is nearly zero, it need not give a small relative error. For $\mathbf k_0\neq\mathbf0$, the leading displacements lie in the one-dimensional subspace spanned by $\mathbf k_0$, and $|k_{0y}\delta_x-k_{0x}\delta_y|\leq\|\mathbf k_0\|\|\mathbf e\|$. This relation avoids dividing by a possibly zero displacement component. If $\mathbf k_0=\mathbf0$, Eq.~\eqref{eq:app_shared_coefficient_bound} still bounds the displacement, but no common direction is defined.

Directional coupling and linear dependence on height are distinct. If $\boldsymbol\beta=\boldsymbol\beta_0$ exactly and calibration is shared, Eq.~\eqref{eq:app_exact_offset} already gives $\boldsymbol\delta=[HZ_0/(Z+H\nu_z)]\mathbf k_0$: both components share one scalar amplitude even without a weak-perspective approximation. Small $H/Z$ and small relative depth variation further make this amplitude approximately $H$. The coefficient is local; the argument does not require one coefficient vector for an entire image or dataset.

The bound also admits approximate calibration sharing. If $\|\mathbf A-\mathbf A_0\|\leq\varepsilon_A$, define $\mathbf k_0=\mathbf A_0\boldsymbol\beta_0/Z_0$ and use $\mathbf A_0$ in Eq.~\eqref{eq:app_shared_coefficient_bound}. The additional residual equals $H(\mathbf A-\mathbf A_0)\boldsymbol\beta/[Z(1+\tau)]$ and is bounded by $H\varepsilon_A(\|\boldsymbol\beta_0\|+\varepsilon_\beta)/[Z_{\min}(1-\eta)]$. Thus small effective-calibration variation contributes an explicit additional error term rather than requiring exact equality.

\subsection{Satellite Evidence and Interpretation of the Width Study}
\label{app:satellite_evidence}

The proprietary IRSA collection used here has a fixed satellite-imaging modality. The related IRSAMap documentation reports that its Chinese imagery was selected from 13 Jilin-1 scenes and standardized to a ground sampling distance of $0.5\,\mathrm m$~\cite{meng_2025_irsamap}. This documents a setting in which many image crops originate from a finite set of acquisitions: additional crops need not supply independently varying camera conditions. These public provenance details are contextual evidence and do not specify the scene inventory or calibration of the proprietary collection.

The ``Jilin-1 GF03D'' technical specifications on the official website of Chang Guang Satellite Technology Co., Ltd. (accessed September 26, 2026) list an orbital altitude of $535\,\mathrm{km}$, pushbroom imaging, and a side-sway range of $\pm45^\circ$. The orbital scale motivates examining $H/Z\ll1$. For illustration, taking an optical depth $Z=535\,\mathrm{km}$ and a building height $H=300\,\mathrm m$ gives $H/Z\approx5.61\times10^{-4}$, so Eq.~\eqref{eq:app_relative_perspective_error} bounds the relative height-perspective correction by approximately $0.056\%$. This is an illustrative calculation, not a measured IRSA bound: orbital altitude is not generally the optical-axis depth $Z$, and the cited specifications do not identify the satellite subtype used for each proprietary image. The pointing range also precludes inferring a common viewing direction across acquisitions from the satellite name alone.

All derivations are conditional on the local pinhole approximation. Spatially varying pushbroom geometry, orthorectification, or mosaicking may change the effective local camera map, and their modeling errors are not included in Eq.~\eqref{eq:app_shared_coefficient_bound}. A common affine image transformation preserves displacement collinearity and maps the residual to the same linear transformation of $\mathbf e$; spatially varying transformations require separate local analysis. In particular, ObliCity reports varying displacement directions within UAV tiles due to low-altitude acquisition and mosaicking~\cite{li_2026_oblicity}, illustrating why one shared coefficient may fail over a larger image region.

The general descriptor retains intrinsic shape, displacement amplitude, and relative geometry as separate factors. Shared calibration and restricted local geometry can reduce their independent variation in a particular acquisition setting, offering a physical interpretation of similar performance at smaller widths. We therefore retain $d=5$ as the default motivated by the general factor organization, without assigning physical identities to its learned channels. These conditional results neither establish a minimum neural width nor identify the cause of non-significant width differences; broader acquisition diversity, rather than dataset openness itself, determines where the local simplifications should be reassessed.

\section{Offset Representation and Readout Parameters}
\label{app:parameter_accounting}

For $K$ instances and offset-token width $d$, the offset representation
contains $Kd$ scalars. The affine readout
$\hat{\mathbf z}=\mathbf W\mathbf o+\mathbf b_{\mathrm{out}}$,
where $\mathbf W\in\mathbb R^{2\times d}$ and
$\mathbf b_{\mathrm{out}}\in\mathbb R^2$, contains $2d+2$
trainable parameters, including the bias.
Table~\ref{tab:app_parameter_accounting} reports these quantities
for selected widths. Counts are per token collection and per
distinct readout module; shared readout parameters are counted
once. These quantities cover the offset representation and its
readout, while the update modules and shared backbone and decoder
contribute additional parameters.

\begin{table}[tp]
\centering
\caption{Offset-representation size and affine-readout parameters,
including bias, for $K=300$.}
\label{tab:app_parameter_accounting}
\begin{tabular}{rcc}
\toprule
$d$ & Token scalars $Kd$ & Readout parameters $2d+2$ \\
\midrule
5   & 1,500  & 12  \\
7   & 2,100  & 16  \\
256 & 76,800 & 514 \\
\bottomrule
\end{tabular}
\end{table}

Relative to $d=256$, $d=5$ reduces token coordinates by
$1-5/256\approx98.05\%$ and readout parameters by
$1-12/514\approx97.67\%$. The reported parameter reduction applies
to each distinct affine readout, including its bias, and does not
describe a full-model parameter reduction.

\end{document}